# A FRAMEWORK FOR NON-MONOTONIC REASONING ABOUT PROBABILISTIC ASSUMPTIONS*

Marvin S. Cohen
Decision Science Consortium, Inc.

## The Problem

Attempts to replicate probabilistic reasoning in expert systems have typically overlooked a critical ingredient of that process. Probabilistic analysis typically requires extensive judgments regarding interdependencies among hypotheses and data, and regarding the appropriateness of various alternative models. The application of such models is often an iterative process, in which the plausibility of the results confirms or disconfirms the validity of assumptions made in building the model. In current expert systems, by contrast, probabilistic information is encapsulated within modular rules (involving, for example, "certainty factors"), and there is no mechanism for reviewing the overall form of the probability argument or the validity of the judgments entering into it.

The present work addresses this problem in the context of conflict resolution in expert systems for image analysis. It involves the design of an expert system inference framework in which probabilistic statements and rules are regarded as assumptions which are explicitly tracked and reevaluated when they lead to conflict among different sources of evidence or lines of reasoning.

Two conceptions of conflict and conflict resolution have been implicit in most approaches to this area. From one point of view, divergence among lines of reasoning can be regarded as stochastic; it is _expected_ to occur some small percentage of the time, due to the chance accumulation of small errors or "noise" in an imperfect process of "measurement". From another point of view, however, divergence can be regarded as a result of faulty beliefs; that is, conflicting results are taken as evidence that one or more premises or forms of argument that led to the conflict are mistaken.

These two conceptions of conflict lead to different rationales for the process of combining evidence or lines of reasoning. From the first point of view, the object is to reduce variance by a blind process of "averaging" akin to that in which chance errors tend to cancel one another out across repeated measurements. From the other point of view, the object is to improve the overall truth of a system of beliefs--to explicitly identify potentially erroneous steps in the argument and to change them.

## A New Approach: Conceptual Outline

The present system, called the Non-Monotonic Probabilist (NMP), regards conflict as jointly knowledge-based and stochastic. It reduces conflict by a process of non-monotonic reasoning prior to statistical aggregation by probabilistic rules. In particular, non-monotonic processes operate on and modify the assumptions and judgments embodied in a belief function model. At the same time, however, the non-monotonic processes are guided by measures of completeness of support provided by the belief function calculus. "Fuzzifying" the latter measures, in turn, ensures a simple but graded process of high level control.

Shaferian belief functions are used as the basic measure of uncertainty, rather than Bayesian probabilities, for several reasons: they do not require numerical definiteness of inputs beyond what the evidence supports; they provide an explicit representation of the quality of an inferential argument;

and they permit "modular" probabilistic analyses based on only subsets of the
evidence. Shafer's system permits a variety of useful specialized models for
representing evidence. One of these special cases is (very nearly) Bayesian
probability theory itself; Shaferian belief functions can represent chance as
Bayesian probabilities do, but permit a simple assessment of the <u>quality</u> or
reliability of those probabilities as well.

The Non-Monotonic Probabilist embeds a belief function model within a super-
structure of non-monotonic reasoning. Non-monotonic logic is a method of
reasoning with incomplete information, in which assumptions may be adopted and
subsequently revised when they lead to contradictory results. Non-monotonic
logic, however, has been <u>exact</u> both in the statements to which it applies and
in its own control mechanisms. As a result, it fails to capture the notion
that support for hypotheses may be <u>graded</u>; and the selection among alternative
equally consistent belief revisions is highly arbitrary. The NMP system ad-
dresses these shortcomings by applying non-monotonic logic to the application
of an uncertainty calculus, and by utilizing measures derived from that cal-
culus to direct the process of belief revision itself.

In the specification of measures for the control of non-monotonic reasoning in
NMP, fuzzy logic has been used. It provides a precise calculus for vague or
imprecise concepts. It thus makes possible the redefinition, in continuous
form, of concepts which occur discretely in traditional non-monotonic systems.
In NMP, for example, "conflict" is a matter of degree, and so is the status of
a statement or rule as an "assumption". As a result, NMP incorporates a
graded control process for belief revision, in which assumptions are subject
to retraction only so long as their resistance to revision is outweighed by
the strength of the conflict.

Applications of NMP to feature identification in aerial photographic images
are being explored. Conflicting results may be obtained from the application
of multiple operators to a pixel array, or from combining extraneous informa-
tion and expectations with one another or with the outcome of a bottom-up
analysis. In these cases, the appropriate response often is a reevaluation of
the reliability of the conflicting sources. Alternatively, their assumed in-
dependence might be questioned, for example by revising the segmentation of
the image; or new analyses might be initiated to confirm hypotheses for which
there is as yet no support, but which could account for the anomaly.

In any probabilistic argument, whether Bayesian or Shaferian, assumptions of
various types are required. Conflict among diverse lines of reasoning can
(and should) force them into the open. To the extent that assumptions are ex-
plicitly tracked and reevaluated, conflict is an occasion for increasing the
validity of a system of beliefs, rather than for meaningless statistical
compromise.

In the following sections, we explore the rationale and the design of NMP in
somewhat more detail. After briefly reviewing the theory of belief functions,
we look at some anomalous results that occur when the belief function calculus
is applied to conflicting evidence. Then we describe some of the NMP process-
es by which such anomalies are eliminated.

<u>Theory of Belief Functions</u>

In the theory of belief functions introduced by Shafer (1976), Bayesian prob-
abilities are replaced by a concept of evidential support. The contrast, ac-
cording to Shafer (1981; Shafer and Tversky, 1983) is between the chance that
a hypothesis is true, on the one hand, and the chance that the evidence <u>means</u>

68

(or proves) that the hypothesis is true, on the other. Thus, we shift focus from truth of a hypothesis to the interpretation of the evidence. As a result, the system (a) is able to provide an explicit measure of quality of evidence, (b) is less prone to require a degree of definiteness in inputs that exceeds the knowledge of the expert, and (c) permits segmentation of reasoning into analyses that depend on independent bodies of evidence.

In Shafer's calculus, support $m(\cdot)$ is allocated not to hypotheses, but to <u>sets</u> of hypotheses. Shafer allows us, therefore, to talk of the support we can place in any subset of the set of all hypotheses. In the case of three hypotheses, $H_1$, $H_2$ and $H_3$, for example, we could allocate support to $H_1$, $H_2$, $H_3$, ($H_1$ or $H_2$), ($H_1$ or $H_3$), ($H_2$ or $H_3$), and ($H_1$ or $H_2$ or $H_3$). As with probability, the total support across these subsets will sum to 1, and each support $m(\cdot)$ will be between 0 and 1. It is natural, then, to say that $m(\cdot)$ gives the probability that what the evidence <u>means</u> is that the truth lies somewhere in the indicated subset.

This device, of allocating support to subsets of hypotheses, enables us to represent the reliability of probability assessments. Suppose, for example, that the presence of a certain type of texture in an image region is associated with a building 70% of the time and with other labels 30% of the time, based on frequency data from a set of training photographs. If we are confident that an image now being analyzed is representative of the training set, we may have m(building) = .7 and m(other) = .3. But if there is reason to doubt the relevance of the frequency data to the present problem (e.g., due to geological or cultural differences between the two geographical areas), we may <u>discount</u> this support function by allocating some percentage of support to the universal set. For example, with a discount rate of 30%, we get m(building) = .49, m(other) = .21, and m({building, other}) = .30. The latter reflects the chance that the frequency data is irrelevant.

Evidence is combined in Shafer's theory by Dempster's rule. The essential intuition is simply that the meaning of the combination of two pieces of evidence is the intersection, or common element, of the two subsets constituting their separate meanings. For example, if evidence $E_1$ means ($H_1$ or $H_2$), and evidence $E_2$ means ($H_2$ or $H_3$), then the combination $E_1 + E_2$ means $H_2$. Since the two pieces of evidence are assumed to be independent, the probability of any given combination of meanings is the product of their separate probabilities. The probability that the combined evidence means any particular subset X is the sum of the products for all combinations that have X as their intersection. When a combination of possible meanings has a null intersection, we know that these meanings cannot co-occur. Thus, the sum of products is normalized to exclude that combination.

### Conflict of Evidence

To what extent does belief function theory yield inferences which are intuitive and plausible in specific applications? A topic of special concern in this regard is conflict of evidence. Zadeh (1984) recently raised an example of the following sort. Suppose we have two experts who we believe to be very reliable and who produce conflicting judgments. For example, there are three possible interpretations of an object x in a specified location: $H_1$--x is a field; $H_2$--x is a forest; $H_3$--x is a building. Analyst A, using photographic evidence, assigns .99 support to $H_1$ and .01 to $H_2$; analyst B, using independent human intelligence information, assigns .99 support to $H_3$ and .01 to $H_2$. We have the following two support functions, and may combine them by Dempster's rule:

69

Table 1

|    | $m_A(\cdot)$ | $m_B(\cdot)$ | $m_{AB}(\cdot)$ |
|----|------|------|------|
| $H_1$ | 0.99 | 0    | 0    |
| $H_2$ | 0.01 | 0.01 | 1.00 |
| $H_3$ | 0    | 0.99 | 0    |

The counterintuitive result, according to Zadeh, is that exclusive support is now assigned to $H_2$, a hypothesis that neither expert regarded as likely. Moreover, the result is independent of the probabilities assigned to $H_1$ or $H_3$.

Shafer's response (in press) is cogent, but ultimately, we feel, off the mark. If we really regard these experts as perfectly reliable, Shafer says, the argument as stated is correct. After all, analyst A says that $H_3$ is impossible, and analyst B rules out $H_1$; that leaves $H_2$ as the only remaining possibility. (Note that exactly the same result is obtained in Bayesian updating if we interpret the $m(\cdot)$ as likelihoods of the evidence given the hypothesis.) On the other hand, Shafer argues that experts are seldom in fact perfectly reliable. A more reasonable procedure would be to "discount" the belief functions supplied by the experts to reflect our doubt in the reliability of their reports. Recalling that we regard these experts as highly reliable (though not perfect), suppose we discount analyst A's belief function by 1% and analyst B's by 2%. The result is the following:

Table 2

|    | $m_A(\cdot)$ | $m_B(\cdot)$ | $m_{AB}(\cdot)$ |
|----|------|------|------|
| $H_1$ | 0.9801 | 0      | .656 |
| $H_2$ | 0.0099 | 0.0098 | .013 |
| $H_3$ | 0      | 0.9702 | .325 |
| $\{H_1, H_2, H_3\}$ | 0.01 | 0.02 | .007 |

We now have a "bimodal" belief function, with the preponderance of support going to $H_1$ and $H_3$. This appears, at first look, to be an intuitively plausible result: it reflects our feeling, which we represented in the form of discount rates, that analyst A <u>or</u> analyst B (or both) could possibly be unreliable.

Note, however, what a vast difference a small amount of discounting makes. In Table 1, after combination by Dempster's rule, there was exclusive support for $H_2$. In Table 2, final support for $H_2$ is only slightly greater than 1%. The second thing to notice is the large discrepancy between $m_{AB}(H_1)$ and $m_{AB}(H_2)$. Although we did in fact discount B at twice the rate as A, the actual numbers (2% and 1%, respectively) and the difference between them were very small. It is by no means clear that the resulting difference in support for $H_1$ and $H_3$ is intuitively plausible. More to the point, the sensitivity of the result for all three hypotheses to very small differences in discount rates is disturbing. Finally, to dramatize the sensitivity even further, note that if support for $\{H_1, H_2, H_3\}$ were 0 for both experts, and if A assigned 0 support to H2, and B assigned 0 support to H2, these very small changes render Dempster's rule indeterminate.

Given the degree of conflict between the two analysts, it seems likely that our original assessment of the reliability was mistaken. Suppose then we discount A by 29% and B by 30%. We now get:

Table 3

|       | $m_A(\cdot)$ | $m_B(\cdot)$ | $m_{AB}(\cdot)$ |
|---|---|---|---|
| $H_1$ | .7029 | 0 | .4243 |
| $H_2$ | .0071 | .007 | .0085 |
| $H_3$ | 0 | .693 | .4044 |
| $(H_1, H_2, H_3)$ | .29 | .30 | .1751 |

Support for $H_1$ and $H_2$ after combination is now roughly equal, certainly a more intuitive result. Then should we have discounted A and B more in the first place? According to Shafer, this is indeed the case; the fault is not in the theory, but in the initial allocation of support.

The example highlights a deeper problem in belief function theory: reliability is to be assessed *as if* we had no knowledge of the evidence actually provided (Shafer in press). We are thus not permitted to *use* the conflict between analyst A and analyst B as a clue regarding their capabilities or as a guide to the appropriate amount of discounting. In ordinary reasoning, however, as just illustrated, it is quite natural to reassess the credibility of an information source in the light of what that source says, or in the light of conflict or corroboration by another source.

There have been efforts to develop Bayesian models that address this issue (e.g., Gardefors, Hansson, and Sahlin, 1983). The essential difference between this work and Shafer's is that the former conditionalizes the assessment of an expert's credibility on what he (and other experts) have actually said (see Shafer, 1984; Cohen, Watson, and Barrett, 1985). This work however, lacks most, if not all, of the virtues of the belief function representation. Formulations which conditionalize on the evidence become extremely complex even for the simplest examples. Little progress has been made in deriving rules for the combination of evidence involving the full range of cases to which Dempster's rule applies. Finally, the pervasive role of prior probabilities in this work is incompatible with the segmentation of evidence into independent arguments which is a significant virtue of Shafer's system.

NMP addresses this problem in a different way, not by revising the belief function calculus, but by supplementing it. It retains the simplicity and modularity of Shafer's representation, while embedding it within a corrective process of qualitative reasoning.

### Fuzzy Control Processes for NMP

NMP deploys a set of fuzzy measures to track and revise assumptions, and generally to guide the application of a belief function calculus. In a certain sense, these measures are *ad hoc*. However, they enable us to avoid an elaborate calculus, like second-order probabilities, which would be computationally unwieldly, and indeed equally *ad hoc*, for this purpose. They provide a graded process of high-level control through a reasonably plausible and simple set of definitions.

Conflict. A simple measure of degree of conflict in a belief function is the following. Let S be a subset of hypotheses and $\bar{S}$ its complement. If $Q = (S, \bar{S})$, then

(1) $$\mu_{conflict}(Q) = 2 \min[Bel(S), Bel(\bar{S})].$$

Where $\mu(\cdot)$ is the degree of membership in a fuzzy set and takes values between

71

0 and 1. This can be justified in two ways. From the fuzzy logic point of view, we might regard it as the membership function for the intersection of belief in S and belief in $\bar{S}$, i.e., a contradiction. Multiplication by 2 normalizes the measure, so that maximum $\mu_{conflict}(Q)=1$ is achieved when $Bel(S) = Bel(\bar{S}) = .5$. Secondly, note that is it equivalent to the following expression:

$$\left\{1 - \frac{|Bel(S)-Bel(\bar{S})|}{Bel(S)+Bel(\bar{S})}\right\}(Bel(S)+Bel(\bar{S})) = 2Bel(\bar{S})$$

when we assume, without loss of generality, that $Bel(S) \geq Bel(\bar{S})$. This expression intuitively captures the notion of conflict in a belief function: the first bracketed expression is the relative similarity of the degrees of belief in S and $\bar{S}$; the larger this is, the greater the conflict. The second bracketed expression is the total <u>committed</u> belief; to the extent that the belief function is "discounted" by assigning support to the universal set $\{S,\bar{S}\}$, we regard the conflict as reduced. In short, the maximum $Bel(S)$ doesn't matter since increasing it (with $Bel(\bar{S})$ constant) has two opposing effects: it increases the difference between $Bel(S)$ and $Bel(\bar{S})$, but also increases the total committed belief.

Conflict resolution is prompted, however, by "significant" conflict, and the degree of significance required may be a variable function of the problem domain. A simple, though somewhat <u>ad hoc</u>, way to accomplish this is to define

$$\mu_{signif.\ conflict}(Q) = \mu_{conflict}^{\gamma}(Q)$$

where $\gamma$ is a power to which $\mu_{conflict}(Q)$ is raised. Increasing $\gamma$ has the effect of requiring higher degrees of conflict to achieve "significance".

<u>Support lists</u>. Dependencies among statements in a standard non-monotonic system (Doyle, 1979) are represented (primarily) by data structures called support lists. A support list justification for a statement has the form (SL <inlist> <outlist>). Such a justification is a valid reason for belief in the statement if every statement in its <u>in</u>list is believed and every statement in its <u>out</u>list is not believed. A non-monotonic justification is a support list whose <u>out</u>list is non-empty. A statement which has a non-monotonic justification is accepted provisionally and may be used in subsequent reasoning: it is rejected when and if some member of its <u>out</u>list comes to be believed.

Similarly, in NMP each rule and each statement is associated with a set of <u>reasons</u>, in the form of a support list. However, in place of a discrete classification (<u>in</u>list vs. <u>out</u>list) we substitute a "fuzzy membership function," i.e., a continuum from <u>in</u> to <u>out</u>. Moreover, it is the current Shaferian <u>support assignment</u> to a statement, rather than the statement itself, which has reasons or which serves as a reason.

Location of a statement $S_i$ on the support list continuum for a second statement $S_j$ or a rule R depends on only two things: (a) the presence of $S_i$ on a list of <u>possible reasons</u> for $S_j$ or R, and (b) the amount of support for the universal set $(S_i,\bar{S}_i)$. In particular, where $S_i$ is a possible reason for $S_j$,

(2a)
$$\mu_{\underline{out}-S_j}(S_i) = m(S_i,\bar{S}_i)$$
$$\mu_{\underline{in}-S_j}(S_i) = 1-m(S_i,\bar{S}_i) = Bel(S_i)+Bel(\bar{S}_i)$$

where <u>in</u> and <u>out</u> hereafter refer to the <u>in</u>list and <u>out</u>list membership functions respectively. Correspondingly, when a rule R is a possible reason for $S_j$,

72

$$\mu_{\underline{out}\text{-}S_j}(R) = m_R(S_j, S_j)$$

(2b)

$$\mu_{\underline{in}\text{-}S_j}(R) = 1 - m_R(S_j, \bar{S}_j)$$

where $m_R(\cdot)$ is the support function assigned by R.

These measures capture a very simple intuition. They place the potential reasons for $S_j$ (or R) in an order corresponding to the relevance of that reason in this particular application, i.e., the reliability or completeness of evidence underlying each reason. To the extent that confidence in $S_j$ or use of R depends upon reasons with high $\mu_{out}$, they rely on unproven (but not disproven) suppositions. (We argue that this is inevitable in any probabilistic analysis.)

What determines the content of the list of possible reasons? For a statement $S_j$, it contains (a) the rules in the system which have a support assignment for $S_j$ in the consequent, and (b) the statements which occur in the antecedents of those rules. The possible reasons for a rule may include a list of potential similarities (or absences of potential dissimilarities) between the target application of the system and the exemplars upon which it was trained. They may also include specifications of model assumptions used to generate support assignments. Finally, they implicitly include assertions of independence of the evidence summarized by the rule from evidence utilized in all other rules of the system.

An elaboration of (2) is motivated by the observation that a statement $S_i$ which is in $S_j$'s support list can have no impact on $S_j$ unless there is a rule linking them (with $S_i$ in the antecedent and a support assignment for $S_j$ in the consequent). Also, of course, a rule R can have no impact on $S_j$ without the (at least partial) satisfaction of its antecedent by a statement. Thus, we must take members of the support list for a statement $S_j$ to be **pairs** of statements and rules $(S_i, R_i)$, rather than statements and rules separately. We now get:

(2')
$$\mu_{\underline{out}\text{-}S_j}(S_i, R) = \min[\mu_{\underline{out}\text{-}S_j}(S_i), \mu_{\underline{out}\text{-}S}(R)]$$
$$= \min[m(S_i, \bar{S}_i), m_R(S_j, \bar{S}_j)]$$
$$\mu_{\underline{in}\text{-}S_j}(S_i, R) = 1 - \mu_{\underline{out}\text{-}S_j}(S, R).$$

A statement or a rule is an assumption to the degree that its acceptance or use depends on what is **possible**, rather than on what is supported by evidence. The following is a simple measure of that concept:

(3)
$$\mu_{assumption}(S_j) = \frac{\sum_{(S,R)} \mu_{\underline{out}\text{-}S_j}(S,R)}{n}$$

where n is the total possible evidence for $S_j$, i.e., the number of statement-rule pairs in the support list for $S_j$. $\mu_{assumption}(S_j)$ is simply the (fuzzy) proportion of $S_j$'s reasons which are **out**, i.e., unsupported by evidence.

Foundations. In standard non-monotonic logic a contradiction in the set of accepted statements triggers a process of dependency-directed backtracking, in which reasons for statements involved in the contradiction are found and revised. A comparable process takes place in NMP. In NMP, however, the process by which beliefs are selected for revision is non-arbitrary. It is guided by measures that reflect both the importance of a reason for a state-

73

ment and the degree to which the reason itself lacks evidential support.

One requirement of dependency-directed backtracking is the ability to find statements or rules which have an indirect impact on a given statement or rule. Suppose a statement-rule pair $(S_i, R)$ is in the support list for $S_j$. Then it has a direct effect on the support assignment for $S_j$ to the extent that $S_i$ or its complement is believed and to the extent that $R$ assigns a non-discounted support function to $S_j$. Other pairs of statements and rules, however, may have an indirect effect on $S_j$ by having an impact on $S_i$ or $R$. All these pairs are, to a degree, part of the "foundations" of $S_j$. We measure this as follows:

(4) $$\mu_{foundations-S_j}(S_k, R_k) = \min_{1 \leq k < j}[\{\mu_{in-S_{k-1}}(S_k, R)\}]$$

In effect, the min function says that the chain of impact linking $(S_k, R_k)$ to $S_j$ <u>via</u> $(S_{j-1}, R_{j-1}) \ldots (S_{k+1}, R_{k+1})$ is only as strong as its weakest link.

To what extent is a statement $S_k$ by itself (or a rule $R$ by itself) part of the foundations of $S_j$? Here, we get:

(5) $$\mu_{foundations-S_j}(S_k) = \sup_R[\mu_{foundations-S_j}(S_k, R)],$$

i.e., $S_k$'s impact is equal to the impact of the most effective chain to which it belongs. Similarly,

$$\mu_{foundations-S_j}(R) = \sup_{S_k}[\mu_{foundations-S_j}(S_k, R)].$$

<u>Suppositions</u>. Suppositions are <u>assumptions</u> with an <u>impact</u>. More precisely, the statements and rules which $S_j$ requires us to "suppose" are
(a) in the foundations of $S_j$, and (b) assumptions in their own right. The degree to which a statement $S_k$ (or a rule $R$) is a supposition of $S_j$ is given by the following:

(6) $$\mu_{supposition-S_j}(S_k) = \min[\mu_{foundations-S_j}(S_k), \mu_{assumption}(S_k)].$$

<u>Dependency-directed backtracking</u>. There are a variety of ways that these measures, or other similar ones, might be used to direct backtracking and belief revision. Here we give one approach. Suppose that $Q = \{S_j, S_j\}$ has a high degree of conflict. The strategy is simply to select the maximal supposition for $S_j$ as the "culprit" $C$, and then to "negate" $C$ by revising the maximal member of $C$'s <u>out</u>list. More precisely, we select a rule or statement $C$ such that
$$\max[\mu_{supposition-S_j}(C')] = \mu_{supposition-S_j}(C).$$
Then we select a statement-rule pair $(S, R)$ for revision such that
$$\max_{S', R'}[\mu_{out-C}(S', R')] = \mu_{out-C}(S, R).$$

Finally, $S$ or $R$ may be revised, depending on which has the least evidential support, i.e., $\max[m(S, \bar{S}), m_R(C, \bar{C})]$.

<u>Conflict as the control over revision</u>. No revisions in fact take place unless the degree of conflict is serious enough to justify them. This involves a simple comparison between the measure of significance of the conflict and a measure of the "resistance" to revision for our best available candidate. Thus, if
$$\mu_{signif.\ conflict}(Q) \geq \mu_{in-C}(S, R),$$

74

S or R may be revised; otherwise, not.

Conclusion

How does NMP relate in general to currently existing AI software tools? Tools for building expert systems now exist which provide for quantitative reasoning about uncertainty (e.g., EMYCIN). Other systems permit qualitative reasoning about and revision of assumptions (e.g., DUCK). NMP is a superset of these capabilities. Our description of it has dwelled on its capability of combining aspects of both: i.e., qualitative reasoning about a quantitative model, and quantitative measures to guide that reasoning. But note that each extreme can be achieved in NMP itself as a special case. If no assumptions are associated with rules or statements, we get a pure system for probabilistic inference (like EMYCIN or PROSPECTOR, with a Shaferian belief function calculus). On the other hand, if all belief functions were to allocate full support between some single hypothesis and the universal set, we get a pure non-monotonic system (like DUCK).

The problems with these extremes are complementary. Pure probabilistic systems never learn anything new about their probabilistic beliefs and assumptions from the experience of applying them. Pure non-monotonic systems do learn, but they have an arbitrariness and an all-or-none quality about the new beliefs they acquire. Our argument, quite simply, is that both capabilities are needed, and that satisfactory systems will, in general, require their combination.

References

Cohen, M.S., Watson, S.R., and Barrett, E. _Alternative theories of inference in expert systems for image analysis_ (Technical Report 85-1). Falls Church, VA: Decision Science Consortium, Inc., January 1985.

Doyle, J. A truth maintenance system. _Artificial Intelligence_, 1979, _12_(3), 231-272.

Gardefors, P., Hansson, B., and Sahlin, N. _Evidentiary value: Philosophical, judicial, and psychological aspects of a theory_. Lund, Sweden: C.W.K. Gleerups, 1983.

Shafer, G. _A mathematical theory of evidence_. Princeton, NJ: Princeton University Press, 1976.

Shafer, G. Jeffrey's rule of conditioning. _Phil. of Sci._, 1981, _48_, 337-362.

Shafer, G. _Probability judgment in artificial intelligence and expert systems_. Lawrence, KS: University of Kansas, School of Business, December 1984.

Shafer, G., and Tversky, A. _Weighing evidence: The design and comparison of probability thought experiments_. Stanford, CA: Stanford University, June 1983.

Zadeh, L.A. Review of Shafer's "A mathematical theory of evidence." _AI Magazine_, 1984, _5_(3), 81-83.

*
This work was supported by the Army Engineering Topographic Laboratories, Contract #DACA72-C-84-0005.